# Marine Engine Fault Dataset: Open-Access Data under Controlled Reference and Fault Scenario Conditions

**Ahmad BahooToroody[1,2], Oleksiy Bondarenko[3], Mohammad Mahdi Abaei [1,2], Niki Yoichi[3], Enrico Zio[4,5]**

1. Department of Energy and Mechanical Engineering, Aalto University, Finland
2. Kotka Maritime Research Centre, Kotka, Finland
3. National Maritime Research Institute, Tokyo, Japan
4. Energy Department, Politecnico di Milano, Via La Masa 34, Milano 20156, Italy
5. MINES Paris-PSL University, CRC, Sophia Antipolis, France

## Abstract

Open-access datasets for marine-engine predictive maintenance remain scarce, particularly those obtained from controlled fault experiments with documented operating conditions, subsystem-level interventions, and system-level measurements. This work presents the Marine Engine Fault Dataset, an openly available experimental dataset generated from a turbocharged, intercooled three-cylinder marine diesel engine operated on a testbed under both reference and fault-scenario conditions. The experimental campaign combined a reference-performance program across the 30–90% load range with scenario-based tests in which abnormal conditions were introduced after stabilized fault-free operation, enabling controlled comparison between baseline and fault-affected behaviour. Five anomaly classes were implemented through physical interventions affecting major engine subsystems: cooling-water pump cavitation, compressor air-filter clogging, air-cooler fouling, injection-valve nozzle clogging and turbine degradation induced through increased exhaust-side restriction. The released data comprise multi-sensor time-series measurements of operating, thermal, pressure, flow and combustion-related variables, together with a separate reference-performance record and accompanying metadata for structured reuse. Technical validation demonstrates that the reference measurements remain physically coherent across the operating range and that the imposed anomalies produce interpretable response patterns consistent with the affected subsystems, including progressively distinguishable behaviour where different anomaly severities were implemented. By combining controlled fault realization, multi-load operation and system-level measurements within a real marine-engine platform, the dataset provides a well-documented benchmark for anomaly detection, fault diagnosis, degradation modelling and related condition-monitoring studies in maritime machinery.

## 1. Background & Summary

Maritime transportation is the backbone of global trade, carrying more than 80% of world trade by volume. However, it is vulnerable to machinery-related failures, which were the leading cause of shipping incidents globally, accounting for more than half of incidents in recent annual records [1]. Although total vessel losses have decreased substantially since the 1990s [2], the reliability of complex shipboard machinery remains a concern. Addressing this concern through data-driven Prognostics and Health Management (PHM) requires representative fault data; however, publicly available real fault datasets from shipboard machinery remain scarce, particularly for marine engines operating under controlled faulty or degrading conditions. This scarcity limits reproducibility, objective benchmarking, and the development of reliable predictive-maintenance methods for maritime systems [3-5].

Predictive maintenance aims to monitor equipment condition and predict failures before they occur, enabling maintenance actions to be performed when needed rather than after failure or according to fixed schedules. Compared with corrective and preventive maintenance strategies, this condition-based approach supports improved reliability, availability, operational efficiency, safety and sustainability [6-13]. Its implementation depends on condition-monitoring systems that record variables such as temperature, pressure, vibration, flow and rotational speed. These measurements support data-driven models, including statistical and machine-learning approaches, for anomaly detection [14], fault diagnosis [15], degradation modeling [16] and remaining useful life prediction [17]. For these models to be developed and evaluated reliably, datasets must include measurements collected under different load conditions, subsystem states and fault conditions, with sufficient documentation of operating context and sensor configuration [18, 19].

However, the availability of publicly accessible datasets suitable for predictive maintenance research is highly uneven across industrial domains. Recent curated overviews identify dozens of publicly available PHM and condition-monitoring datasets; for example, a 2025 PHM dataset library reports 59 datasets, and a broader 2026 overview lists more than 90 publicly available datasets [20, 21]. These datasets are primarily concentrated in a limited number of application areas, including bearings [22, 23], batteries [24, 25], turbofan engines [26, 27], manufacturing processes [28, 29], and PHM challenge datasets [26, 30-32], mostly within aerospace, manufacturing, and parts of the energy sector. In contrast, comparable datasets for maritime machinery, particularly ship main engines under faulty or degrading operating conditions, remain limited [20].

This has led researchers in the maritime domain to rely on simulation-based datasets, surrogate benchmarks or proprietary operational measurements that are not publicly accessible, limiting reproducibility and cross-study comparison [27, 32-35]. For example, in a recent study on marine gas turbines [36] algorithm validation has relied on datasets such as NASA's C-MAPSS turbofan dataset [27, 32] and the marine propulsion simulation system (MPSS) dataset [37], underlining also that available data for marine gas turbines remain limited and that C-MAPSS differs substantially from marine machinery in usage, operating conditions and boundary conditions. This issue has significantly influenced how data-driven PHM methods are developed and validated in the maritime domain, where the lack of widely accepted benchmark datasets makes objective comparison between different methods more difficult. This limited availability of publicly accessible maritime machinery fault datasets is driven not only by the rarity of failure events, but also by fragmented data ownership across ship owners, operators, equipment manufacturers and service providers, as well as commercial confidentiality, liability exposure, cybersecurity concerns and the operational difficulty of collecting well-annotated fault data under actual sea-going conditions [33-36].

Beyond limited accessibility, the suitability of existing datasets for PHM research is constrained by technical and experimental factors. Several review studies on machinery prognostics indicate that datasets suitable for degradation modeling and remaining useful life prediction should include run-to-failure trajectories, fault progression information and consistent fault labeling; however, such characteristics are rarely available in many industrial applications, including maritime systems [18, 38, 39]. Moreover, most publicly available datasets are developed for individual components rather than complete engineered systems, and therefore, do not capture system-level interactions such as subsystem coupling, load sharing and operational dependencies [18, 40, 41]. In addition, some datasets (e.g., the IMS bearing dataset [42], the FEMTO/PRONOSTIA bearing dataset [30] and the NASA C-MAPSS turbofan engine dataset [27, 32]) are typically collected under controlled operating conditions and represent simplified or single-fault scenarios, which limits their applicability to machinery operating under variable load and environmental conditions typical of marine engines. Consequently, datasets that combine realistic operating conditions, multiple fault scenarios, documented degradation progression, and multi-sensor measurements at the system level remain limited, constraining systematic benchmarking and reproducible evaluation of predictive maintenance algorithms in the maritime domain [20-23].

To address these limitations, this study presents the Marine Engine Fault Dataset, an openly available dataset collected from a marine diesel engine testbed under representative operating conditions, including both healthy and faulty scenarios with documented fault progression and multi-sensor time-series measurements. The dataset includes multiple fault types affecting different subsystems of the engine, including the cooling system, air intake system, combustion system and turbocharger, enabling system-level investigation rather than analysis of isolated components. The experiments were conducted under multiple engine load conditions and a separate test campaign was performed to construct a reference condition map of the engine across a wide range of operating points using a design-of-experiments (DOE) approach. This reference map supports the development of baseline performance models and enables model-based and hybrid prognostics approaches.

Fault development was implemented in controlled stages to represent degradation progression, enabling the dataset to be used for research on anomaly detection, fault diagnosis and degradation modeling. By providing open access to the dataset along with detailed documentation of operating conditions, fault scenarios and sensor measurements, this work supports reproducible research and systematic benchmarking of predictive maintenance methods in maritime and related industrial applications. The dataset is not intended to replace field operational data, but rather to provide a controlled, well-documented experimental benchmark that enables systematic method development and comparison under realistic marine engine operating conditions.

## 2. Method

### 2.1. Experimental setup

The Marine Engine Fault dataset was generated from a controlled experimental campaign conducted on a marine engine test platform at the National Institute of Maritime, Port and Aviation Technology (MPAT), National Maritime Research Institute (NMRI). Measurements were acquired under controlled reference and fault-scenario tests using a marine diesel engine test setup.

The test platform was based on a Matsui iron works MU323DGSC marine engine coupled to a water brake dynamometer for controlled loading.

The engine is a turbocharged intercooled, four-stroke unit operating on Marine Diesel Oil (MDO) / Heavy Fuel Oil (HFO) fuel. Its rated power is 257 kW and its rated speed is 420 rpm. The engine has a 230 mm bore, a 380 mm stroke and three cylinders. The main engine specifications are summarized in Table 1.

Table 1. Test engine specification

| **Parameter** | **Value** |
| --- | --- |
| Model | Matsui iron works MU323DGSC |
| Type | Turbocharged intercooling |
| Cycle | 4-stroke |
| Fuel | MDO / HFO |
| Rated power | 257 kW |
| Rated speed | 420 rpm |
| Bore | 230 mm |
| Stroke | 380 mm |
| Number of cylinders | 3 |

During all tests, the engine was operated along the propeller load characteristics, which were reproduced using a water brake dynamometer. The engine operating points were set to follow the propeller law, with engine speed regulated by a governor to maintain the target value at each operating point and torque imposed by the load device. Each test stage was conducted at fixed operating conditions. While the engine can be run with both MDO and HFO, all tests were conducted using the same fuel type, in particular MDO, to ensure consistency across reference and faulty operating conditions. Ambient conditions, primarily air temperature, were recorded throughout the experiment. Steady-state conditions were considered to be achieved once the main monitored parameters (e.g., engine speed, torque, and key temperatures and pressures) remained within predefined tolerance bands for a specified duration. A stabilization period of at least 15 min was applied before data acquisition at each operating point. In addition, each anomaly test was preceded by approximately 1 h of fault-free steady-state operation to ensure a stable baseline prior to fault introduction.

The operating conditions used in the experimental campaign are summarized in Table 2, including the load levels and test durations applied for the reference performance measurements and for each implemented anomaly scenario.

Table 2. Operating conditions used in the experimental campaign

| **Test category** | **Component / system** | **Anomaly / condition** | **Engine load (%)** | **Test duration per load (h)** |
|---|---|---|---|---|
| Fault-scenario test | Cooling water system | Pump cavitation | 60, 85 | 2 |
| | Maine Engine (ME) air intake system | Air filter clogging | 40, 60, 75, 85 | 2 |
| | ME compressor | Air cooler fouling | 40, 60, 75, 85 | 2 |
| | ME combustion system | Injection valve nozzle clogging (one hole) | 40, 60, 85 | 1 |
| | | Injection valve nozzle clogging (two holes) | 40 ~ 75 | 3 |
| | ME turbocharger | Turbocharger degradation | 40, 60, 85 | 2 |
| Reference-performance test | ME general | Reference performance | 30 ~ 90 | 0.25 |

### 2.2. Instrumentation and Data Acquisition

The overall configuration of the engine test bed and measurement system is shown in Fig. 1. The abbreviations used in the engine test-bed schematic are provided in Appendix A, Table A1. The measured variables, including their units and sensor locations, are summarized in Appendix A, Tables A2 and A3, and the instrumentation and data-acquisition specifications are provided in Appendix A, Table A4.

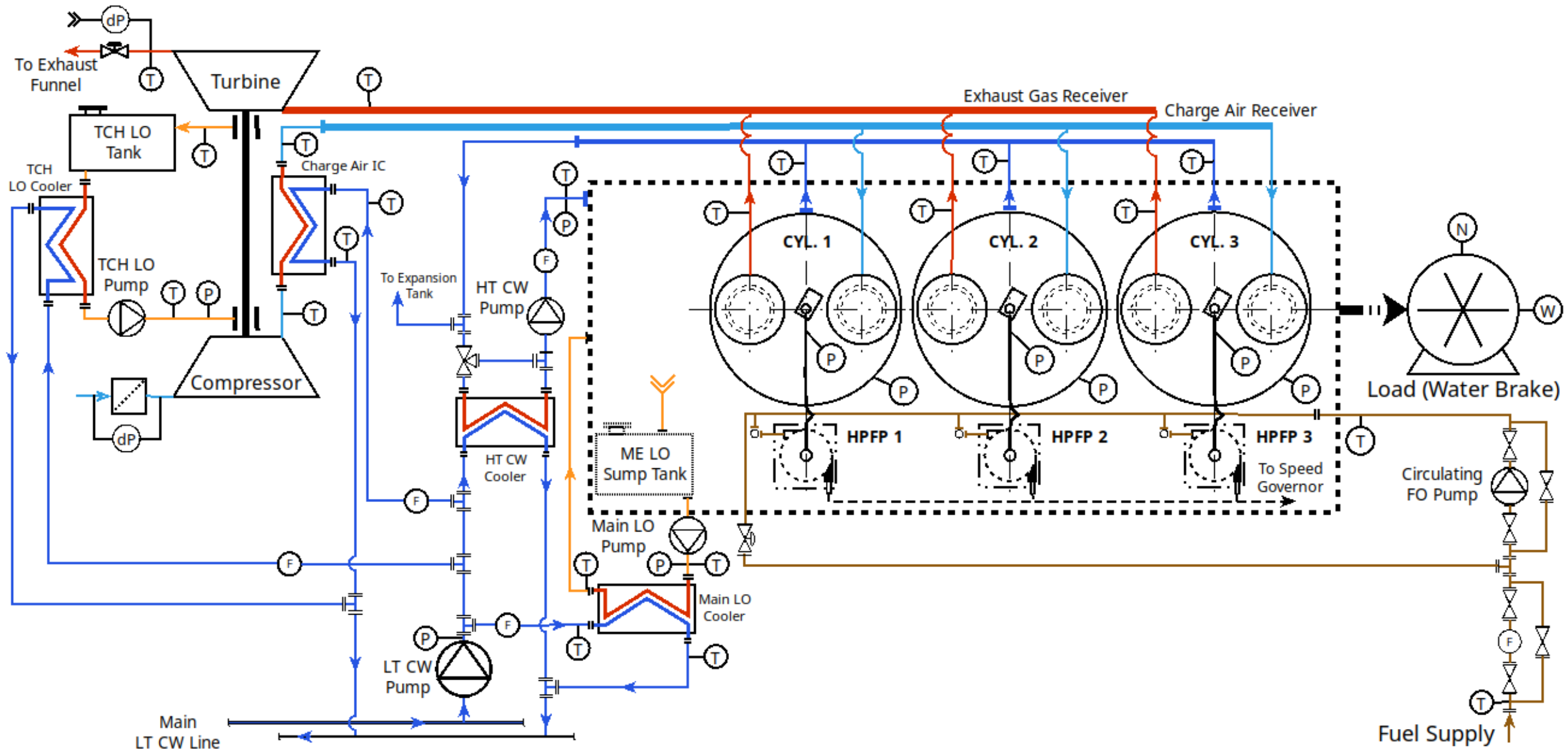

Fig. 1. Schematic of the engine test bed showing the main components and instrumentation used for data acquisition

Temperature measurements were performed using K-type thermocouples, wheras pressure measurements were obtained using strain-gauge-based sensors. The pressure measurements were acquired using Kyowa PAV-10KF and water-cooled Kyowa PEF-S-20MPSA1 sensors. Low-frequency data, including temperature and low-pressure signals, were acquired using a Keithley 2700 data acquisition system operating in scanning mode, with a typical scan interval of approximately 1~2 sec. High-frequency measurements were recorded using a Contec AI-1204Z-PCI data acquisition card, which continuously acquired the raw in-cylinder pressure signal on a crank-angle basis (measurement session is driven by a trigger aligned with the piston top dead center (TDC) position). Detailed specifications of the sensors and acquisition systems are summarized in Table 6.

The two acquisition systems operated concurrently but independently. Synchronization was achieved at the data-processing stage by associating each set of derived high-frequency metrics with the corresponding timestamp of the low-frequency scan. The measured variables included shaft speed, water-brake load, pressure and differential-pressure signals, flow rates, and temperature measurements across the engine, cooling-water, lubricating-oil, intercooler, fuel-oil, and turbocharger subsystems. Their locations and signal definitions are provided in Tables 4 and

5. In addition, the high-speed data acquisition system provided crank-angle-resolved in-cylinder pressure measurements, from which key combustion and performance metrics were calculated and subsequently merged into the dataset at the corresponding low-frequency sampling instants.

## 2.3. Experimental design

The experimental design comprised two main parts: reference-performance measurements under normal operating conditions and scenario-based tests under abnormal conditions. Each experimental scenario consisted of two sequential phases: an initial steady-state operation under fault-free conditions, followed by a gradual introduction of the fault condition. A key feature of the design was that failure development was represented as a continuous exponentially growing process and, then, discretized into fixed stages for practical execution during the engine tests.

The reference-performance campaign was carried out over the 30%–90% load range by operating the engine at constant loads and speeds. The operating points were selected according to a Design of Experiment (DOE) plan and engine performance was measured at these conditions; the distribution of the measured operating points on the engine characteristic plane is visualized in Fig. 2.

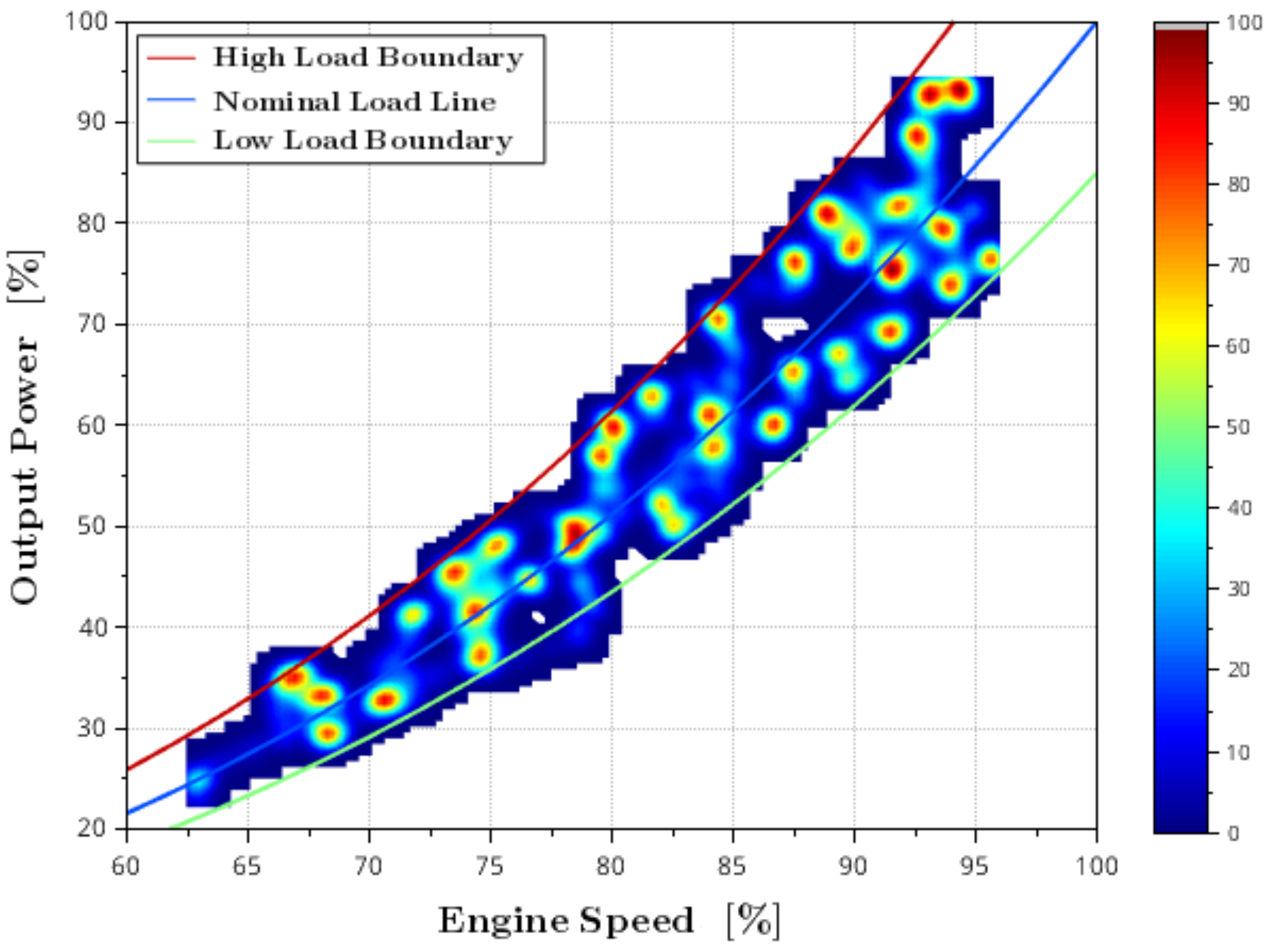

Fig. 2. Distribution of experimentally measured operating points on the engine characteristic plane following the DOE plan for the reference-performance campaign.

To represent the system-level relation between the implemented fault scenarios and overall engine performance degradation, a fault tree analysis (FTA) was developed a posteriori. The FTA links the basic events corresponding to the implemented scenarios to the top event of engine performance degradation and organizes the major pathways into three main subsystem branches: the cooling water system, the fuel injection system and the turbocharging system. In this way, the FTA provides a structured representation of how the selected fault scenarios relate to subsystem-level functional degradation and to the overall loss of engine performance. The experimentally implemented fault routes are highlighted in red in Fig. 3.

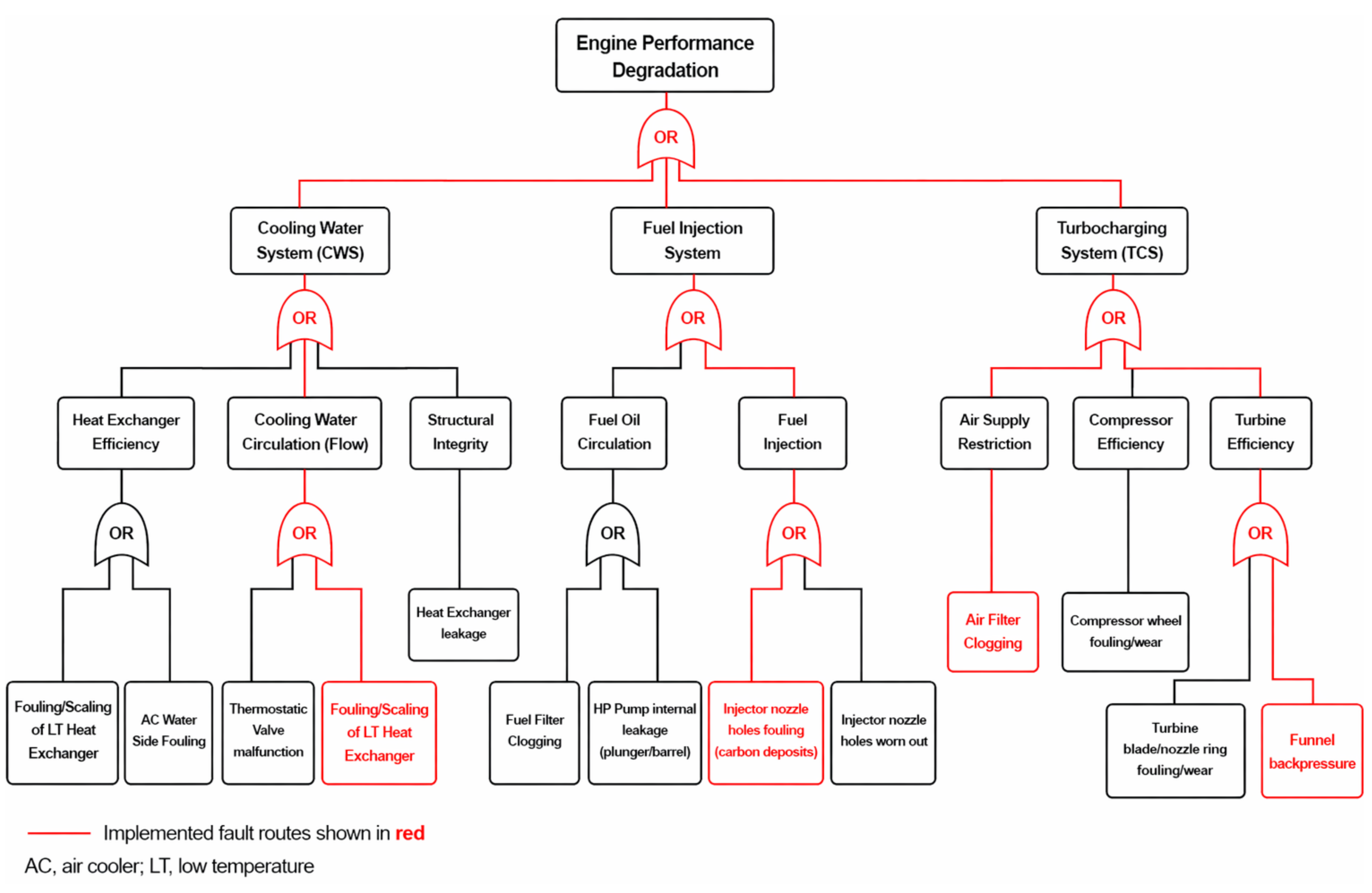

Fig. 3. Fault tree analysis (FTA) linking subsystem-level degradation pathways to the top event of engine performance degradation.

The experimentally implemented fault routes are highlighted in red. The implemented experiments were based on a set of failure scenarios organized by engine component or subsystem. These scenarios covered the cooling water system, air intake system, compressor-side air cooler, combustion system and turbocharger-related exhaust path. The final set of implemented scenarios included pump cavitation, air filter clogging, air cooler clogging, injection valve nozzle clogging and turbocharger degradation. The physical implementation of these scenarios is described in Section 2.4

### 2.4 Fault Scenario Implementation

The fault scenarios were realized through controlled physical interventions on the test engine. The nominal test conditions for the reference-performance and fault-scenario measurements are summarized in Table 2. The implemented interventions included air injection into the cooling-water pump, partial restriction of the air filter, alteration of the cooling-water flow through the air cooler, partial blockage of injector nozzle holes by welding and partial closure of the exhaust-stack valve to increase back pressure after the turbine.

Air-cooler degradation, notably water side fouling-like anomaly, was introduced by changing the charge-air temperature setpoint at different engine loads. This intervention produced progressive changes in the charge-air and cooling-water temperatures. At low loads and temperatures above 44 °C, hunting was observed in the charge-air temperature because the cooling-water flow control valve operated near its dead-band.

Air-filter degradation, notably filter material clogging-like anomaly, was introduced by gradually covering the filter surface with tape in order to restrict the air supply. This intervention increased the pressure drop across the filter and affected turbocharger operation.

Injection-valve nozzle clogging was implemented using specially prepared nozzles with one hole and two holes plugged by welding, representing different anomaly severities. Representative photographs of the prepared injector assemblies and the corresponding faulty atomizers are shown in Fig. 4. The one-hole-plugged nozzle was used for tests under abnormal operating conditions at different engine loads, whereas the two-hole-plugged nozzle was tested under a dynamic load program to represent more severe and varying degradation.

Pump cavitation was imitated by injecting pressurized air into the suction side of the cooling-water pump. This approach reproduced the symptoms of cavitation without causing permanent destructive damage to the test equipment. The injected air bubbles mimicked two-phase-flow conditions, and the resulting measurements showed a pressure drop together with signal instability in the cooling-water circuit.

Turbocharger degradation was implemented by using a shut-off valve installed on the exhaust stack. Partial closure of this valve increased exhaust back pressure, which reduced the turbine expansion ratio and mimicking a decrease in turbine efficiency.

Taken together, these interventions converted selected degradation mechanisms in major engine subsystems into experimentally implemented fault scenarios for structured dataset generation.

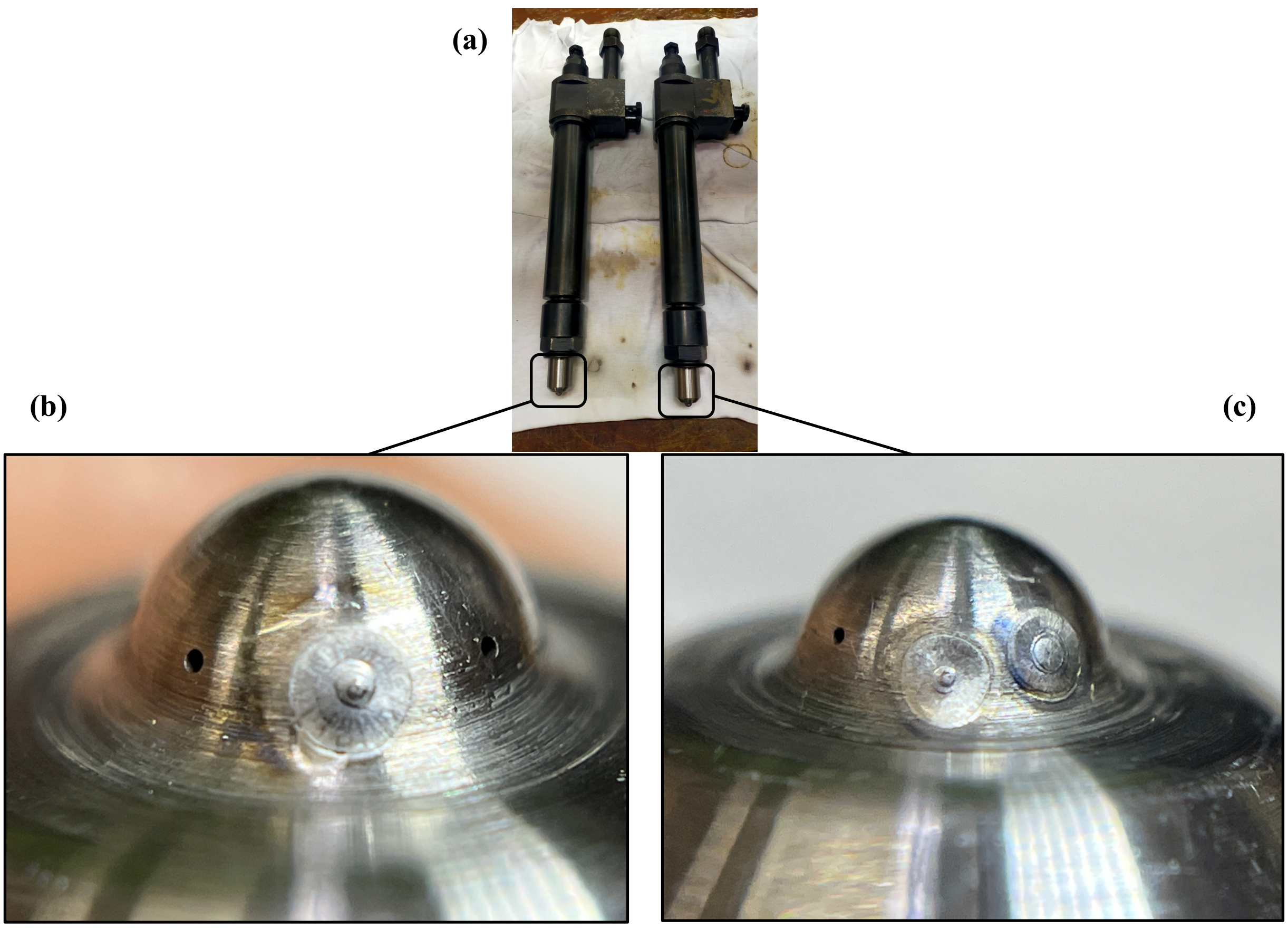

Fig. 4. Representative photographs of the physical interventions used to implement selected fault scenarios: (a) injector assemblies prepared for the injection-valve nozzle clogging scenario; (b) atomizer with one plugged nozzle hole, representing the lower-severity intervention; and (c) atomizer with two plugged nozzle holes, representing the higher-severity intervention.

## 3. Data records

The Marine Engine Fault Dataset is publicly available at **10.5281/zenodo.19857425** and contains the reference-performance and fault-scenario measurements acquired from the marine engine test platform described in Section 2. As shown in Fig. 5, the released package is organized into five scenario-specific folders; AC Clogging, AC Fouling, Injector Nozzle, Pump Cavitation and Turbine Degradation, together with a baseline reference file (Reference_Data.csv) and three supporting metadata files: README_public_release.md, variable_dictionary.csv, and dataset_index.xlsx.

The scenario folders contain the CSV records corresponding to the implemented fault scenarios and tested operating conditions. File names identify the scenario and load condition directly, enabling straightforward retrieval of the relevant measurements. In general, the scenario files contain time-related fields together with the recorded engine variables, and the header rows provide the full variable names, shorthand labels, and units. The anomaly records are therefore distributed as scenario-specific files rather than as a single merged table, preserving the experimental structure of the campaign.

The fault-free baseline measurements are provided separately in Reference_Data.csv. This file should be treated as the reference record for the campaign rather than assumed to be fully column-compatible with all scenario files. The distinction between the reference file and the anomaly files is therefore intentional and reflects their different experimental roles.

To support reuse, the repository also includes file-level and variable-level metadata. The dataset_index.xlsx file provides a file inventory of the released records and their scenario assignments, whereas variable_dictionary.csv documents the released signals and indicates whether each variable appears in the reference file and/or the scenario files. The README_public_release.md file provides a repository-level description of the package structure, naming convention and practical guidance for interpretation. Together, these files provide the documentation needed to navigate and reuse the dataset across the five implemented fault scenarios.

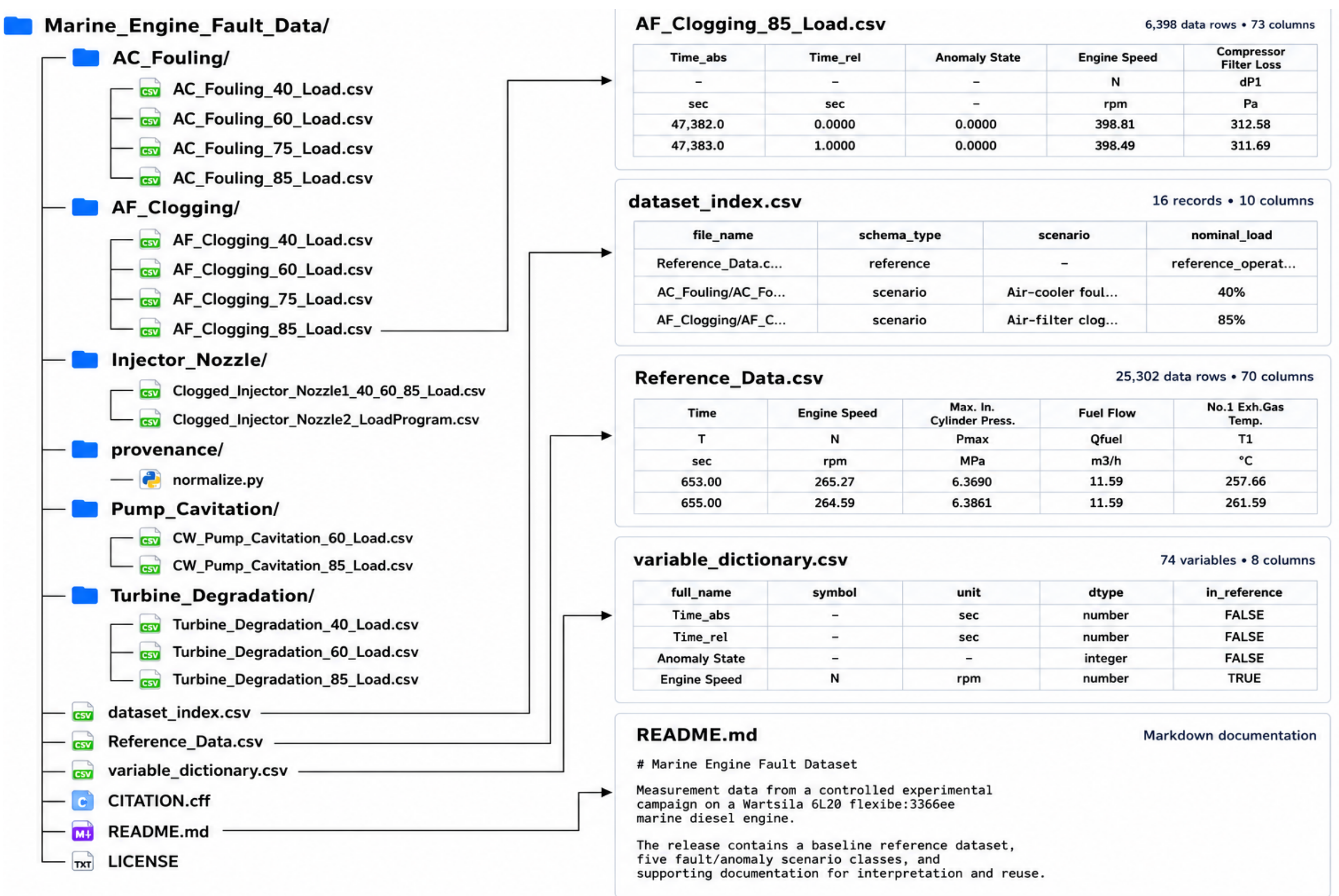

Fig. 5. Directory structure and representative file previews of the released marine engine fault dataset. The preview panels illustrate representative contents of a scenario file, the reference file, the variable dictionary, the dataset index, and the README file.

## 4. Technical Validation

Technical validation was carried out to confirm that the released records are consistent with the controlled operating conditions of the experimental campaign and that the implemented fault scenarios generate physically meaningful and reproducible deviations from the reference-performance measurements. The validation, therefore, addressed three main questions: whether the reference-performance data exhibit coherent behaviour across the defined operating range, whether the recorded signal changes under each fault scenario are plausible in relation to the affected subsystem, and whether the staged implementation of degradation is reflected by progressively distinguishable patterns in the measurements. Accordingly, this section evaluates the dataset as a reliable experimental resource for anomaly detection, degradation modelling and related condition-monitoring studies, rather than as a benchmark of predictive model performance.

### 4.1. Reference-performance consistency

The reference-performance records provide the baseline for interpreting deviations observed under the implemented fault scenarios. This subsection examines whether selected variables show coherent behaviour across the reference-performance operating range and whether the corresponding steady-state records remain sufficiently stable for later comparison with fault-scenario measurements.

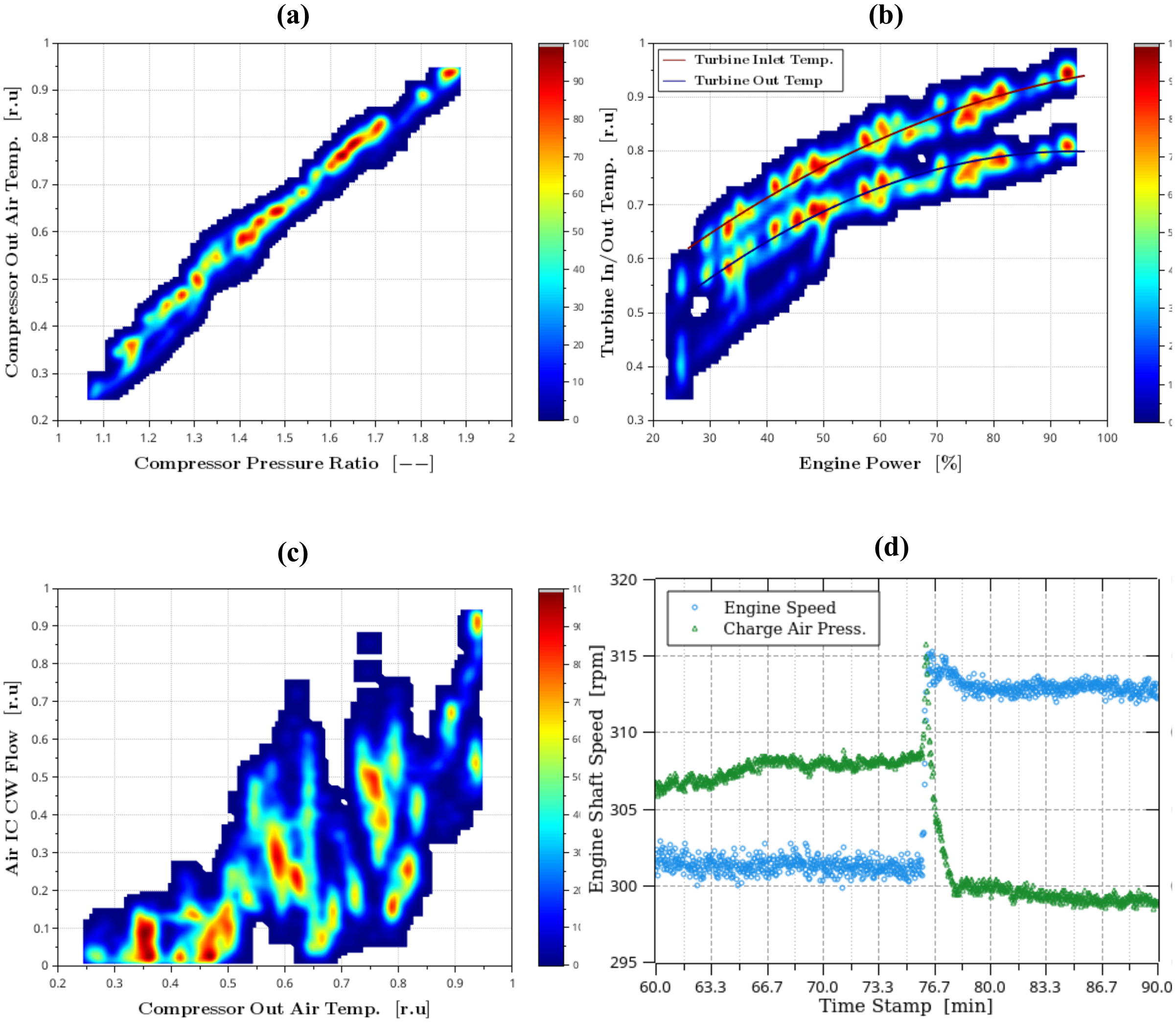

Fig. 6. Consistency of the reference-performance dataset across representative variables. (a) Compressor outlet air temperature vs. compressor pressure ratio. (b) Turbine inlet and outlet

temperatures vs. engine load. (c) Intercooler cooling water flow vs. compressor outlet air temperature. (d) Time series of engine shaft speed and scavenging air pressure during steady-state operation and load transition.

The consistency of the reference-performance dataset is illustrated in Fig. 6 throughout a set of representative variables covering the operating, thermal and pressure/flow variables across loads. Panel (a) shows the relationship between compressor outlet air temperature and compressor pressure ratio, where a clear monotonic trend is observed across the entire operating range. The low scatter around the main trajectory indicates stable compressor behaviour and consistency of the air-side thermodynamic response. Panel (b) presents turbine inlet and outlet temperatures as a function of engine load. Both variables exhibit physically consistent trends, with increasing load associated with higher turbine inlet temperature and a corresponding rise in exhaust outlet temperature, indicating stable combustion and energy conversion characteristics. Panel (c) demonstrates the dependence of intercooler cooling water flow on compressor outlet temperature. While a general increasing trend is evident, the data exhibit a relatively large dispersion. This behaviour is attributed to the discrete (impulse-type) operation of the cooling water control system, which results in frequent valve actuation and stepwise adjustments of the flow rate. Importantly, despite this variability, the overall trend remains consistent with the expected thermal response, indicating that the observed scatter reflects control characteristics rather than measurement instability. Finally, panel (d) shows a representative time history of engine shaft speed together with scavenging air pressure at two operating points, including a transition between them. The steady-state segments exhibit stable levels and negligible fluctuations in both variables, confirming stable operation and reliable attainment of target conditions.

Overall, the examined variables exhibit physically consistent and interpretable behaviour across the operating range. The primary thermodynamic relationships remain well-defined and monotonic. The short segment of steady-state time history further confirm that the operating points are reproducible and sufficiently stable for analysis. Taken together, these observations indicate that the reference-performance dataset is internally coherent and provides a reliable baseline for subsequent comparison with fault-induced deviations.

### 4.2. Scenario-response plausibility

Fault-scenario records should exhibit response patterns that are physically consistent with the subsystem affected by each implemented anomaly. The scenario-specific causal trees in this subsection summarize the dominant response pathways expected from the imposed interventions and identify the variables most relevant for validation. Unlike the system-level FTA in Section 2.3, they are used here only to support interpretation of the measured anomaly responses. For each fault scenario described in Section 2.4, the causal tree defines the expected physics-based propagation pathway, which is then examined against representative experimental observations to assess whether the measured response follows the anticipated behavior.

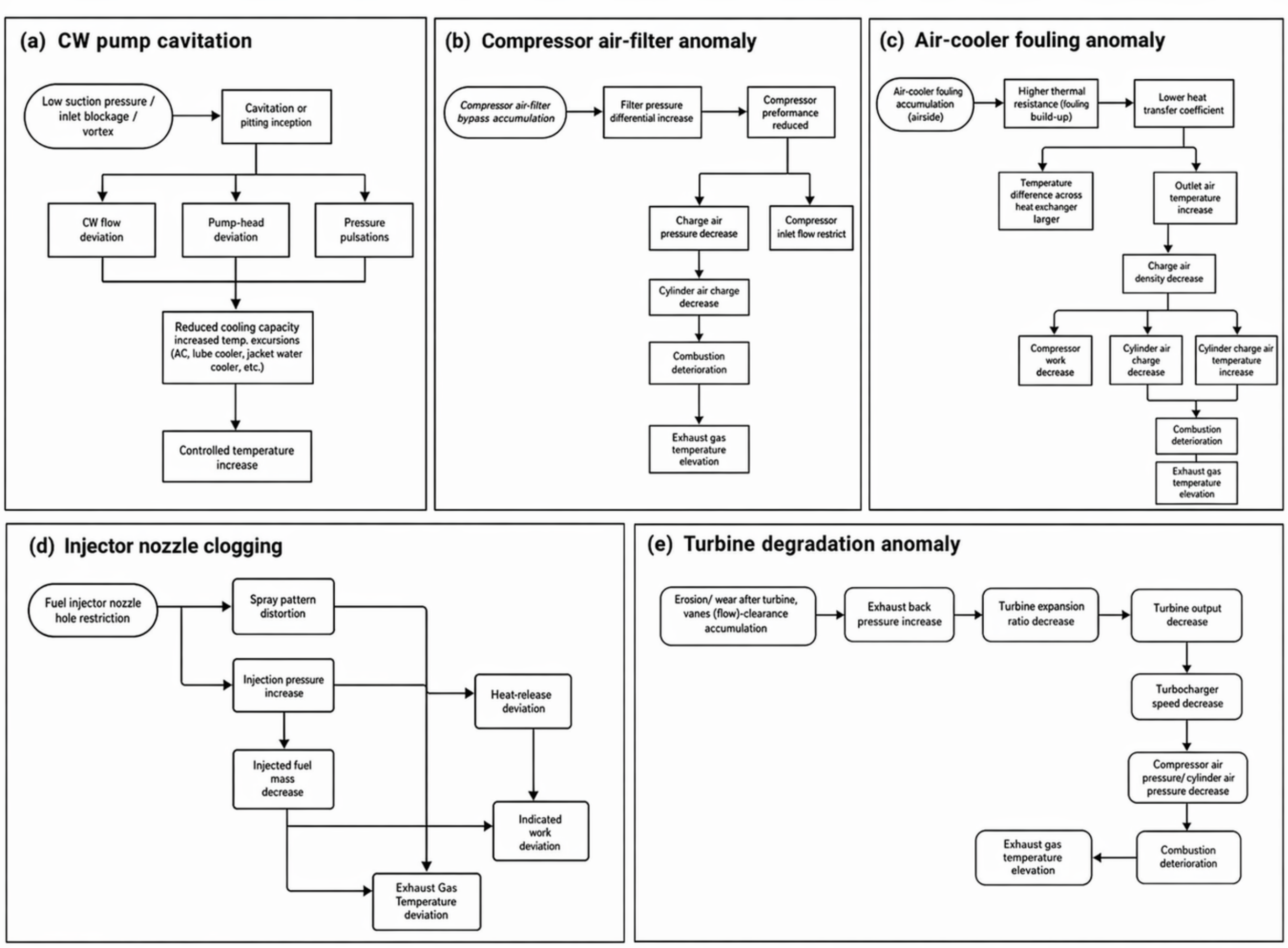

Fig. 7. Scenario-specific causal trees for the implemented anomaly cases: (a) Cooling-Water (CW) pump cavitation; (b) compressor air-filter clogging; (c) air-cooler fouling; (d) injection-valve nozzle clogging; and (e) turbine degradation anomaly. The diagrams summarize the expected physical response pathways for interpreting the measured anomaly responses.

Cavitation at the pump impeller, shown in Fig. 7(a), acts as the primary hydraulic disturbance in the cooling-water circuit, leading to reduced effective flow, loss of pump head and pressure pulsations. These effects are expected to propagate through the circuit primarily via hydraulic pathways, with the most immediate and reliable indicators being changes in flow behaviour and unsteady pressure characteristics at downstream of the pump. Although a reduction in cooling effectiveness may occur, the associated temperature response depends on the severity of the disturbance and the thermal inertia of the connected systems and may therefore be weak or not directly observable under mild cavitation conditions. Accordingly, the most informative variables for this scenario are those related to cooling-water flow and pump-related pressure dynamics, with temperature-based indicators considered secondary and conditional. To support the casual interpretation, Fig. 8 represents a representative time history of the pressure measured at the pump outlet using an auxiliary sensor whose measurements are not included in the main dataset. The signal clearly exhibits pressure drop at air injection onset followed by pressure pulsations associated with cavitation, providing direct evidence of the underlying hydraulic disturbance.

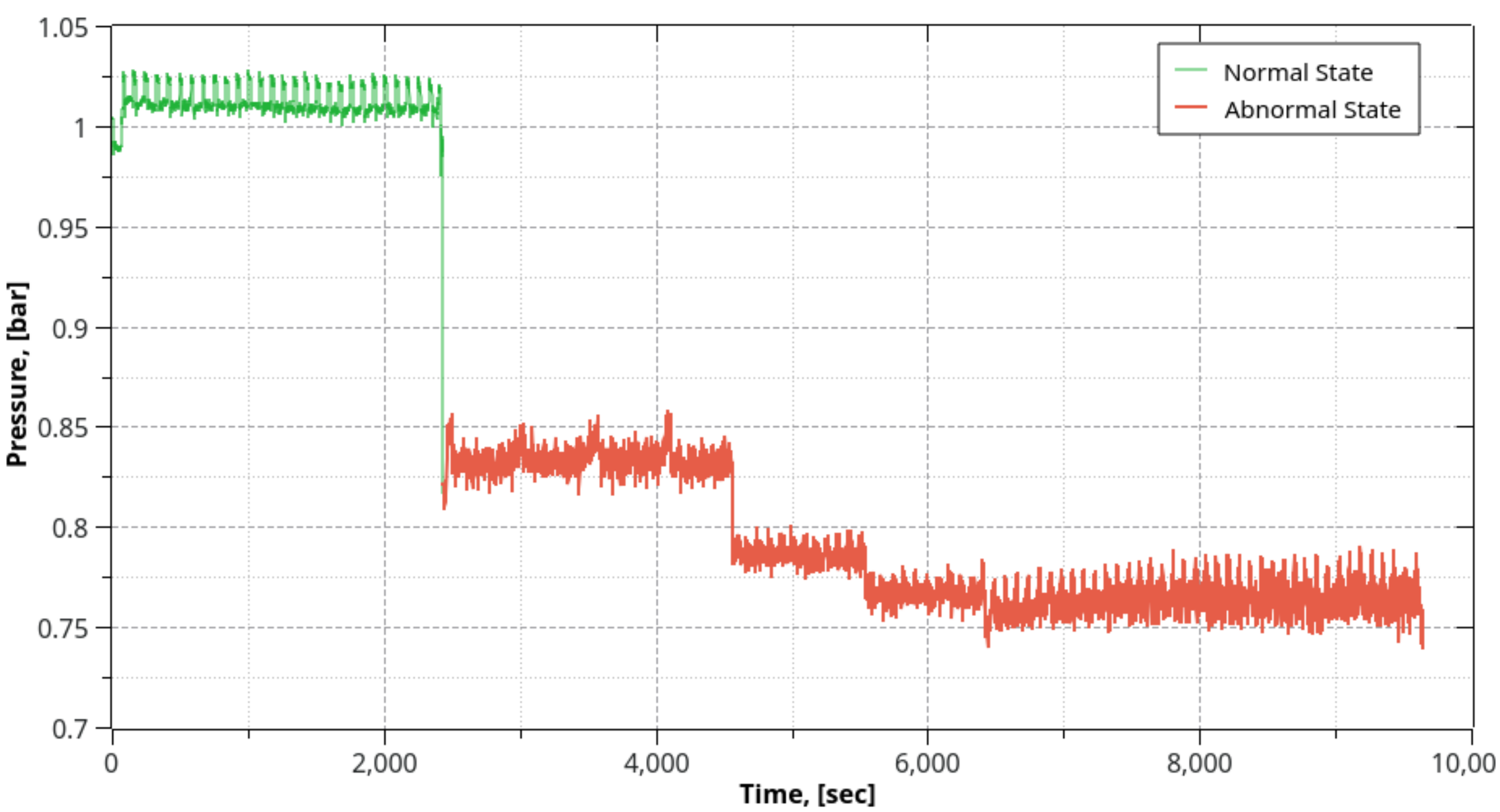

Fig. 8. Pressure trend at CW pump outlet during imitation of cavitation-like anomaly

In Fig. 7(b), the anomaly pathway begins with an increase in filter pressure difference and proceeds through reduced compressor pressure ratio to lower cylinder air charge. Accordingly, the response should be visible not only in the local filter-related variable, but also in charge-air pressure and downstream combustion-sensitive quantities. The key validation question is whether intake restriction is reflected in reduced air supply and, ultimately, in deviation of exhaust-gas temperature, the extent of which depends on the operating conditions and engine-turbocharging matching.

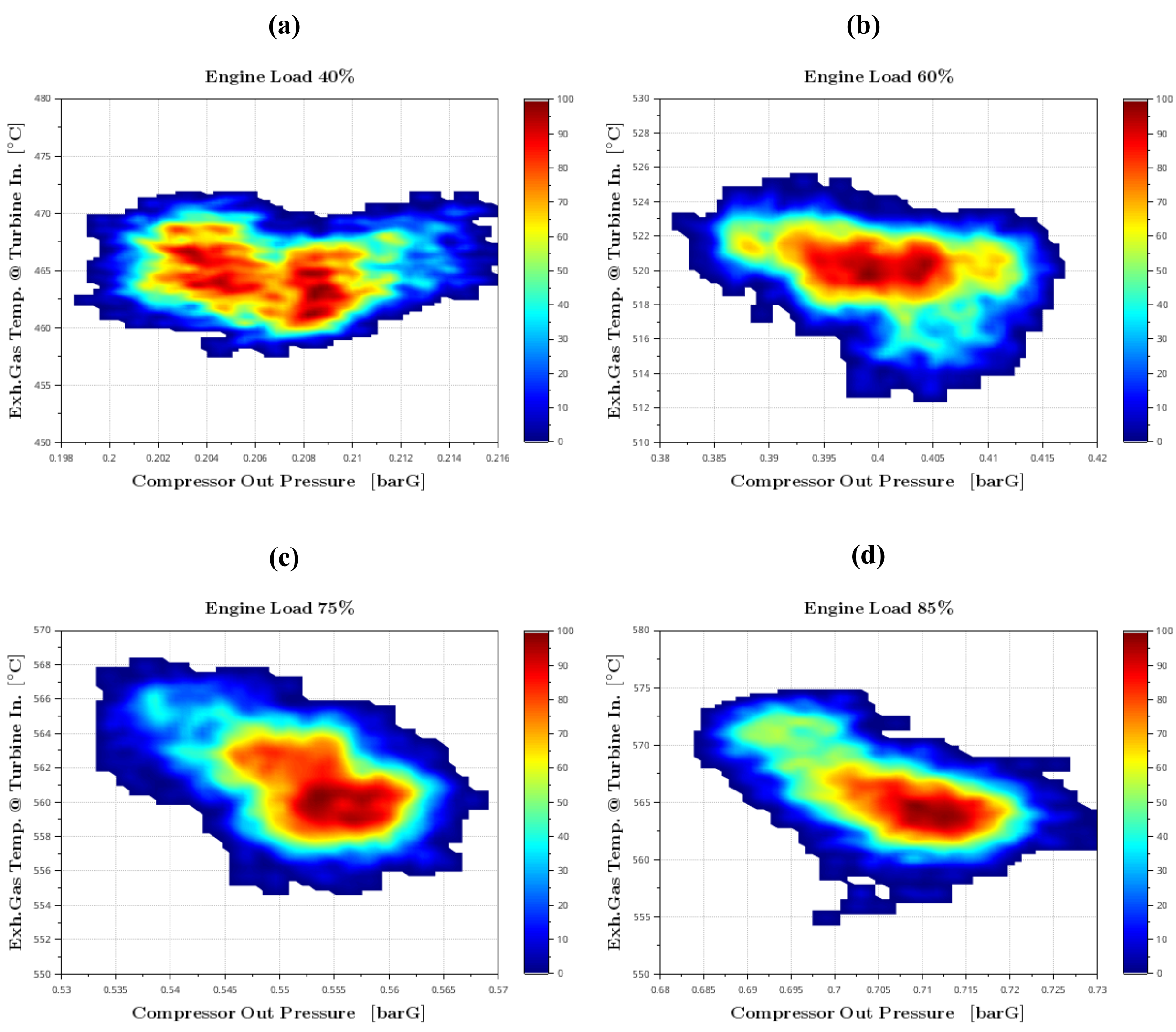

Fig. 9. Turbine inlet temperature versus compressor outlet pressure at different engine load conditions (40% (a), 60% (b), 75% (c), and 85%(d)).

To examine this behaviour, Fig. 9 presents the relationship between turbine inlet temperature and compressor outlet pressure at several load conditions. As can be seen, at low loads, the trend is weak, indicating that moderate intake restriction has a limited impact on combustion temperature under these conditions. As the load increases (75-85%), a clear inverse relationship emerges, with reduced compressor pressure corresponding to elevated exhaust gas temperature. This behaviour is consistent with increasing sensitivity of combustion to air supply.

As shown in Fig. 7(c), impaired heat transfer, rather than direct air-path blockage, is the governing mechanism in the air-cooler fouling scenario. The expected response is an increase in outlet air temperature together with altered temperature differences across the heat exchanger, whereas the airflow itself remains largely unaffected. This leads to a reduction in charge-air density and, consequently, degraded cylinder charging conditions. The most relevant variables for this scenario are therefore those describing the heat-exchanger performance and charge-air thermodynamic state, with exhaust-gas temperature serving as a downstream indicator of combustion deterioration.

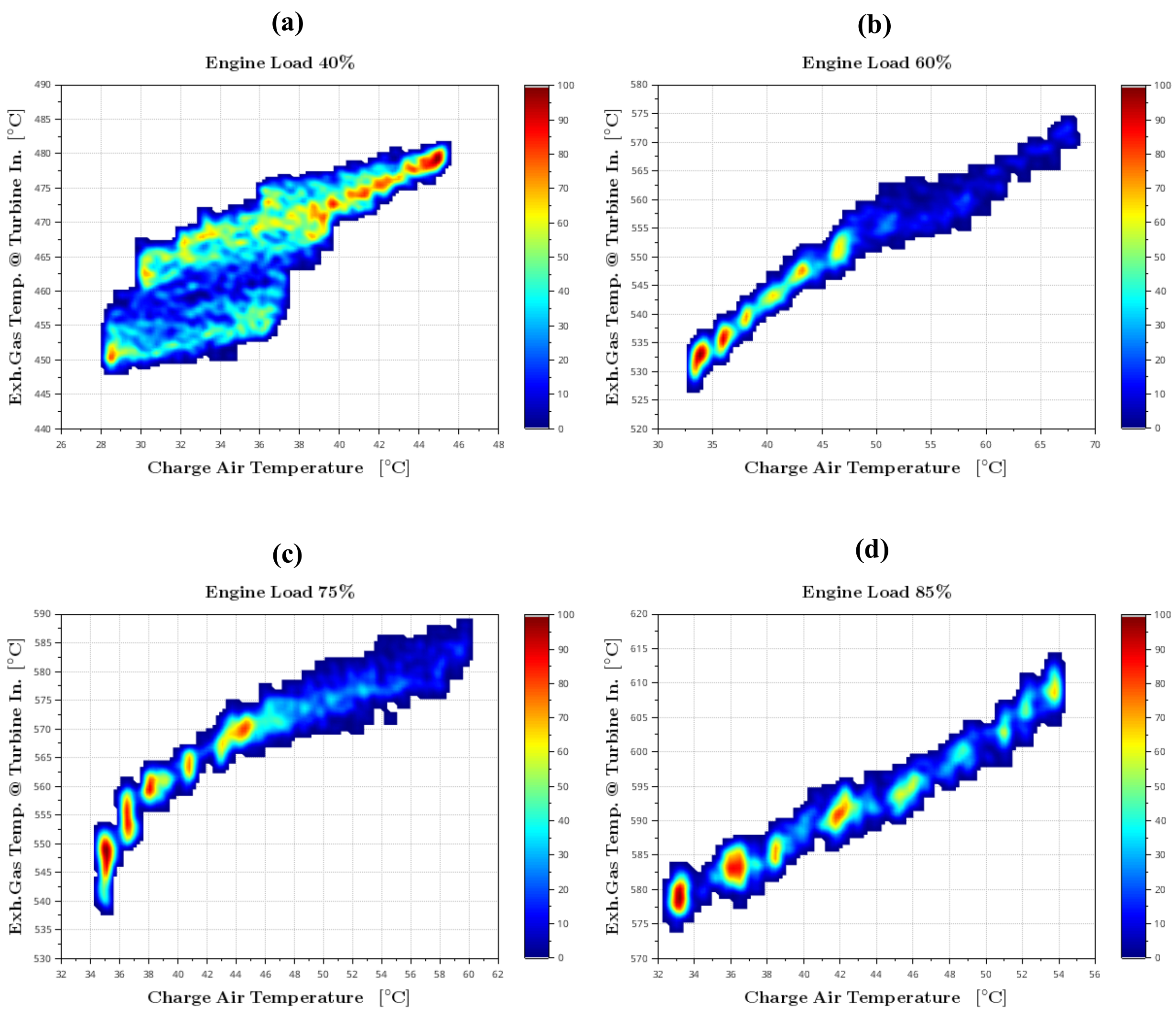

Fig. 10. Turbine inlet temperature versus charge air temperature across different engine load conditions (40% (a), 60% (b), 75% (c), and 85%(d)).

This behaviour is illustrated in Fig. 10, which shows the relationship between charge air temperature and turbine inlet temperature across the considered load conditions. In contrast to the intake restriction scenario, clear and consistent positive correlations are observed at all loads. The robustness of these trends across the operating range confirms that the dominant effect of the air-cooler is thermal in nature, with a direct and monotonic propagation from reduced heat-transfer effectiveness to combustion response.

Fig. 7(d) shows that nozzle-hole restriction affects the system primarily through degradation of the fuel injection and combustion process, rather than through a directly observable hydraulic signature. Partial blockage of the nozzle holes reduces the effective flow area and distorts the spray formation process, leading to impaired atomization and altered local fuel-air mixing in the affected cylinder. Although injection pressure may also be influenced, the most informative observable consequences are changes in cylinder-resolved combustion behaviour. Accordingly, the response of this anomaly is evaluated using variables such as maximum cylinder pressure and indicated work, whereas exhaust-gas temperature is considered a secondary integrated indicator of the resulting combustion imbalance.

The response was examined under multiple operating conditions and for different levels of nozzle restriction in order to assess both the detectability and the progression of the anomaly. For the one-hole-plugged case, the response is illustrated through the distribution of maximum combustion pressure across the three cylinders at several load conditions, as shown in Fig. 11(a-c). The affected cylinder (No. 3) consistently exhibits lower peak pressure than the healthy cylinders, indicating reduced effective heat release. Although the separation between the distributions remains modest, it is sufficiently clear to confirm a persistent cylinder-specific combustion imbalance under the lower-severity intervention.

For the more severe two-hole-plugged case, the effect is examined under a stepwise load increase. Fig. 11(d) presents the distribution density of maximum combustion pressure as a function of engine load. A clear and persistent deviation of the affected cylinder from the normal cylinders is observed across the operating range, and the magnitude of this deviation increases with load. This behaviour is consistent with the increasing sensitivity of combustion to spray quality and fuel distribution at higher fueling rates, and confirms that the more severe nozzle restriction produces a stronger and more sustained abnormal response.

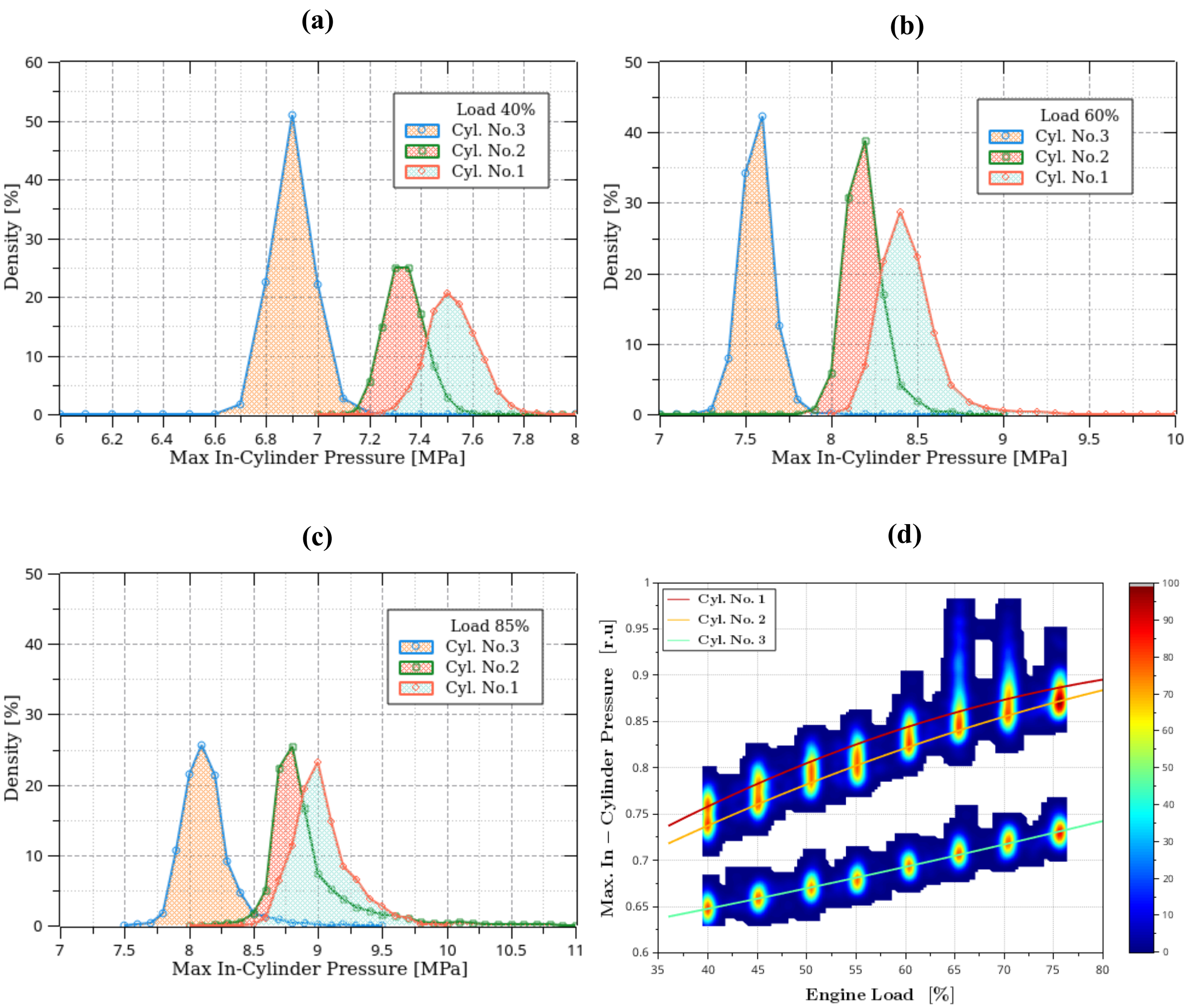

Fig. 11. Cylinder-resolved maximum combustion-pressure response under the injection-valve nozzle-clogging anomaly. Distribution densities of maximum in-cylinder pressure for the one-hole-plugged case at fixed loads of (a) 40%, (b) 60%, and (c) 85%, and distribution density of maximum in-cylinder pressure versus engine load for the two-hole-plugged case under a stepwise load increase (d).

The atomization photographs provide additional physical support for this interpretation. As shown in Fig. 12, the two modified atomizers produced visibly different spray patterns: the two-hole-plugged case exhibited fewer visible plumes and a more asymmetric spray distribution than the one-hole-plugged case. This qualitative evidence is consistent with more severe atomization degradation and supports the stronger signal-level deviations observed in the measured combustion variables.

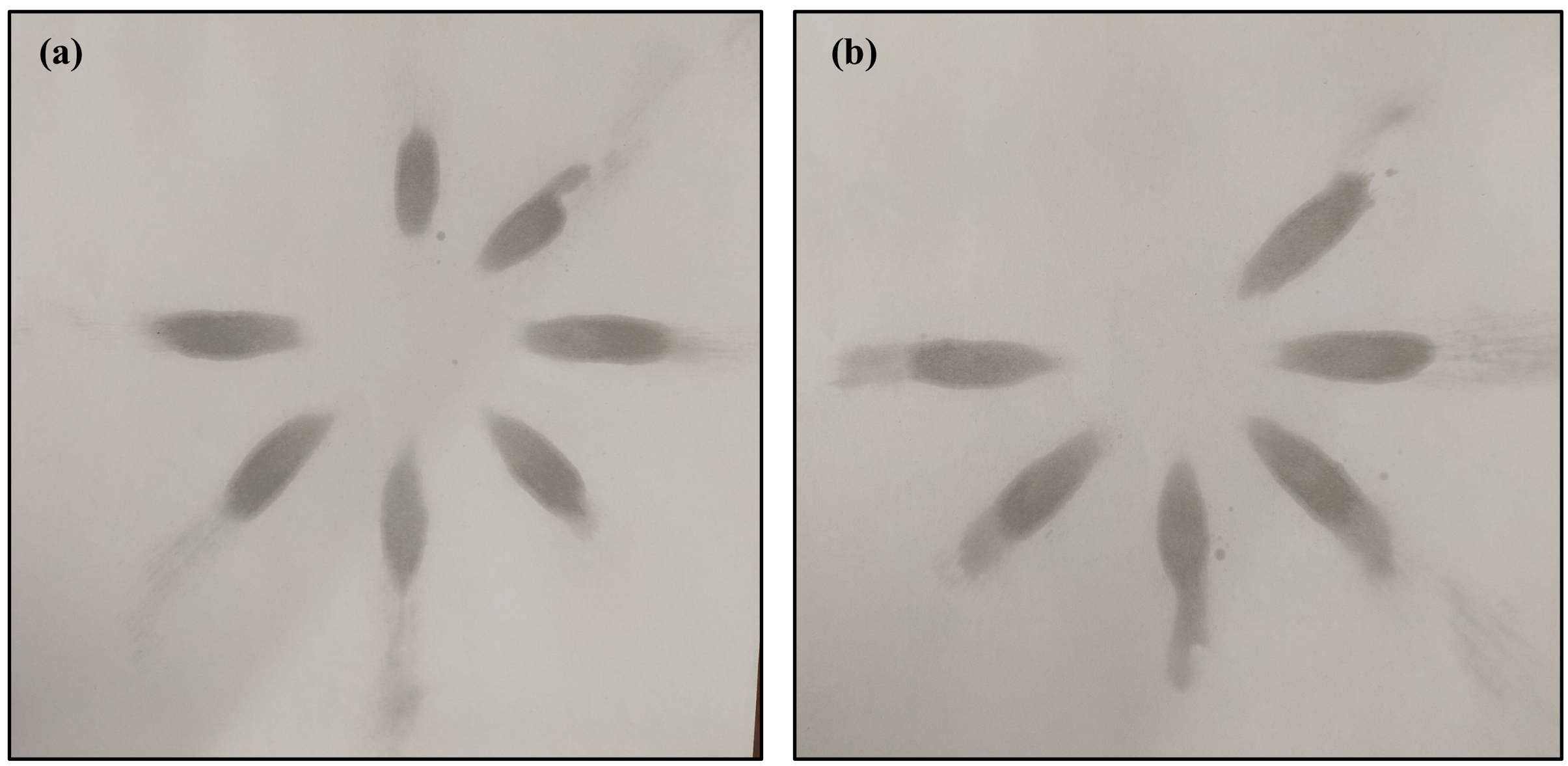

Fig. 12. Representative atomization patterns of the modified injector atomizers used in the injection-valve nozzle clogging scenario: (a) atomization shape obtained with one plugged nozzle hole; and (b) atomization shape obtained with two plugged nozzle holes.

In Fig. 7(e), turbine degradation is represented experimentally through controlled increase of exhaust backpressure after the turbine, which reduces the effective expansion ratio and, consequently, the turbine power available to drive the compressor. This approach emulates the loss of energy extraction capability associated with degraded turbine performance. The resulting disturbance propagates through the turbocharging system, leading to reduced compressor pressure ratio and altered air-supply conditions, with subsequent impact on combustion-related variables such as exhaust-gas temperature. Accordingly, the most relevant measurements for this scenario include exhaust-side pressure, charge-air pressure and representative exhaust-temperature channels, which together capture the coupled response of the turbine-compressor system.

To further characterize this behaviour, a turbocharger energy conversion ratio is introduced as a system-level indicator reflecting the balance between turbine energy extraction and compressor work. In the present study, this quantity is approximated using temperature differences across the turbine and compressor, and is defined as:

$$\eta_e^* = \frac{T_{c,out} - T_{c,in}}{T_{t,in} - T_{t,out}} \quad (1)$$

where $T_{c,in}$ and $T_{c,out}$ denote the compressor inlet and outlet temperatures (here it should be noted that $T_{c,in}$ corresponds to engine room temperature in the data set). $T_{c,in}$ and $T_{c,out}$ denote the turbine inlet and outlet temperatures. Furthermore, the temperature differences are normalized with corresponding inlet temperatures at the reference (fault-free) condition. This formulation provides a dimensionless indicator of the relative magnitude of compressor-side temperature rise to turbine-side temperature drop, and is used here to assess changes in the effective energy transfer within the turbocharger system.

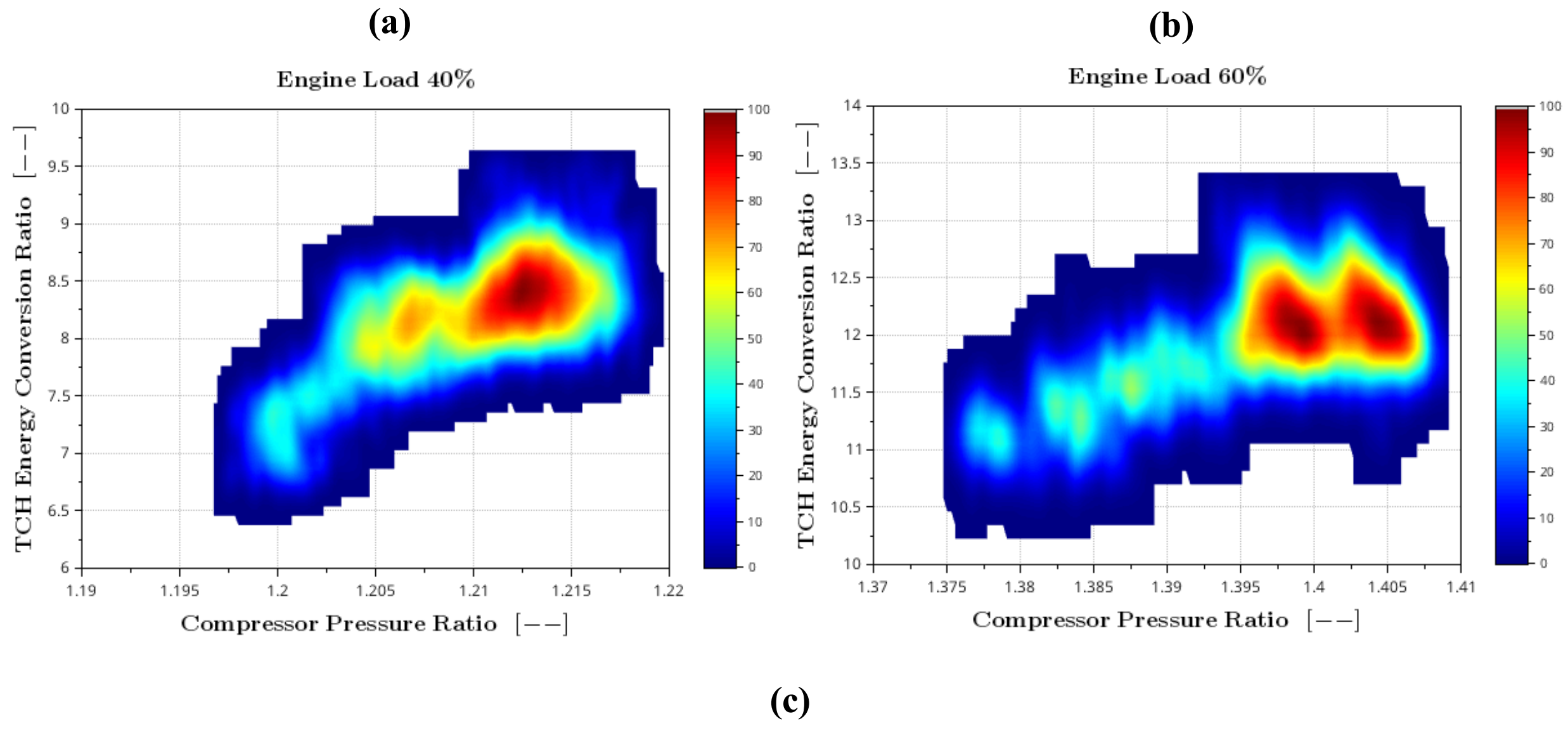

(c)

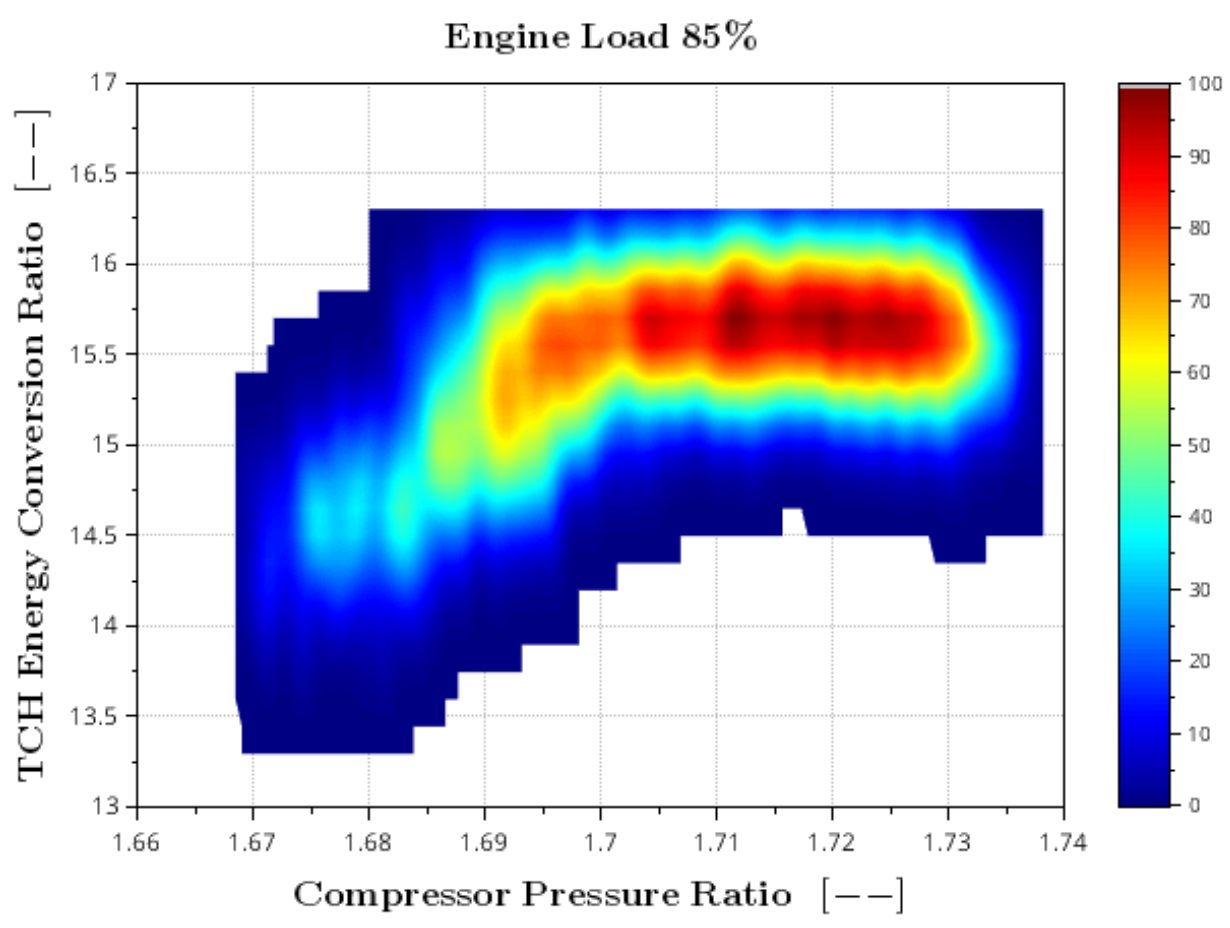

Fig. 13. Turbocharger energy conversion ratio versus compressor pressure ratio under the turbine degradation anomaly at engine loads of (a) 40%, (b) 60%, and (c) 85%.

Fig. 13 presents this ratio as a function of compressor pressure ratio. A distinct nonlinear behaviour, at high loads, is observed, with the ratio remaining approximately constant over a range of operating conditions and, then, exhibiting a rapid decline at lower pressure ratios. This transition indicates a breakdown of effective energy transfer within the turbocharger system under increased exhaust restriction, consistent with reduced turbine expansion capability and off-design operation of both turbine and compressor.

## 5. Usage Notes

This dataset is most appropriately used for scenario-wise analysis rather than as a single pooled Table. Before normalization, feature extraction or between-file comparison, users are encouraged to identify the target scenario, operating condition and experimental structure of each file.

Comparisons with the fault-free baseline are best carried out only after variable correspondence and operating-condition compatibility have been verified. In particular, the released files are not all structured for identical preprocessing workflows. The injector-nozzle records may require particular care, because they represent dedicated restricted-nozzle test programs rather than the same fixed-load structure used in several other scenarios.

The *Anomaly State* field should be interpreted at the file level, not as a globally uniform label across the full release. Likewise, shorthand variable labels should not be used alone for automated channel matching where ambiguity may arise; variable mapping should rely on the full variable-name row and the accompanying variable dictionary. Additional care is required for pressure-related channels provided in volts, which represent raw sensor signals rather than converted engineering pressure values.

For reproducible reuse, automated parsing and cross-scenario analysis should be performed only after consultation of the dataset index and variable dictionary.

## 6. Code Availability

No custom code is associated with the released dataset. The records are provided in CSV format together with accompanying documentation, a dataset index, and a variable dictionary to support interpretation and reuse.

## 7. Data Availability

The dataset described in this study is publicly available at **10.5281/zenodo.19857425** under the **CC4.0**. The released package includes the reference-performance file, the scenario-specific fault records, and the accompanying metadata and documentation required for interpretation and reuse.

## 8. Acknowledgements

This work was supported by the Research Council of Finland under the Academy Research Fellowship project TRANSITION-MODEL [grant no. 362947].

## 9. Author Contributions

Ahmad BahooToroody, Mohammad Mahdi Abaei and Enrico Zio conceived the study and developed the experimental concept underlying the dataset. Ahmad BahooToroody and Oleksiy Bondarenko discussed and finalized the fault scenarios. Oleksiy Bondarenko and Niki Yoichi designed and prepared the experiments, led the data collection. Oleksiy Bondarenko cleaned and technically validated the data, and prepared the associated Figures. Ahmad BahooToroody reviewed the released data, prepared the metadata and utility code. Ahmad BahooToroody co-drafted the manuscript with Oleksiy Bondarenko. All authors revised and approved the final manuscript.

## 10. Competing Interests

The authors declare no competing interests.

## 11. Appendix A. Test-bed nomenclature, measured variables, and instrumentation

Table A1. Abbreviations used in the engine test-bed schematic.

| Symbol | Description |
|---|---|
| P | Pressure measurement points |
| T | Temperature measurement points |
| F | Flow measurement points |
| dP | Pressure difference measurement points |
| h | Relative position measurement |
| N | Engine shaft speed measurement |
| W | Water brake weight measurement |
| HT | High Temperature |
| LT | Low Temperature |
| LO | Lubricating Oil |
| CW | Cooling Water |
| TCH | Turbocharger |
| HPFP | High Pressure Fuel Pump |
| CYL | Cylinder |
| IC | Intercooler |
| FO | Fuel Oil |

Table A2. Summary of operating, pressure, and flow-related variables, units, and sensor locations.

| Signal ID | Measured quantity /Unit | Location/Description |
|---|---|---|
| N | rpm (revolution per minute) | Water brake shaft open end / engine shaft rotational speed |
| F | kgf | Water brake / reaction force of the water brake |
| Pturb | kgf/cm$^2$ | Turbocharger's compressor outlet / pressure developed by the compressor |
| Qfuel | $m^3$/h | After fuel supply pump and before fuel circulating loop / fuel consumption by the engine |
| dPf | Pa | Compressor air filter / pressure drop across filter |
| dPex | Pa | Turbocharger's turbine outlet / pressure difference with respect to atmospheric |
| Pl_lo | Raw signal in Volts (V) | Main LO pump / pump delivery pressure |
| Pl_water1 | V | HT CW pump / pump delivery pressure |
| Pl_water2 | V | LT CW pump / pump delivery pressure |
| Pl_loturb | V | TCH LO Pump / pump delivery pressure |
| Qw_eng | m3/h | HT CW loop / volume flow of cooling water through the engine |
| Qw_lo | m3/h | Main LO CW loop / volume flow of cooling water through Main LO cooler |
| Qw_air | m3/h | IC CW loop / volume flow of cooling water through IC |
| Qw_turb | m3/h | TCH LO CW loop / volume flow of cooling water through TCH LO cooler |

Table A3. Summary of temperature variables, units, and sensor locations.

| Signal ID | Measured quantity /Unit | Location/Description |
|---|---|---|
| T1, T2, T3 | °C | Cylinder exhaust outlet pipe / exhaust gas temperature after cylinder |
| T4 | °C | Turbocharger's turbine inlet / exhaust gas temperature at turbine inlet |
| T5 | °C | Turbocharger's turbine outlet / exhaust gas temperature at turbine outlet |
| T6 | °C | Engine cylinder block / HT Cooling water temperature at engine inlet |
| T7, T8, T9 | °C | Cylinder cooling loop outlet / Cooling water temperature at engine cylinders outlet |
| T10 | °C | Engine cylinder block / LO temperature at engine inlet |
| T11 | °C | Engine cylinder block / LO temperature at engine outlet |
| T12 | °C | Main LO cooler / Cooling water temperature at cooler inlet |
| T13 | °C | Main LO cooler / Cooling water temperature at cooler outlet |
| T14 | °C | Charge Air IC / Intercooler inlet air temperature |
| T15 | °C | Charge Air IC / Intercooler outlet air temperature |
| T16 | °C | Charge Air IC / inlet CW temperature |
| T17 | °C | Charge Air IC / outlet CW temperature |
| T18 | °C | Fuel oil circulating loop / FO temperature at engine inlet |
| T19 | °C | Turbocharger / inlet LO temperature |
| T20 | °C | Turbocharger / outlet LO temperature |
| T21 | °C | FO flow meter / inlet FO temperature |
| T22 | °C | Temperature in the room |

Table A4. Instrumentation and data-acquisition specifications.

| Sensor/System | Measured Quantity | Range/Accuracy | Application |
|---|---|---|---|
| K-Type Chromel-Alumel thermocouple | Temperature | -200°C~1200°C ±0.75 % RO | General temperature measurements |

| Kyowa PAV-10KF | Pressure | 10 bar<br>±0.36 % RO | General pressure measurements |
|---|---|---|---|
| Kyowa PEF-S-20MPSA1 | Pressure | 20 MPa<br>±0.39 % RO | High pressure in a high temperature environment |
| NAGANO KEIKI GC62 | Pressure difference | 0 ~ 1 kPa, 0 ~ 10 kPa<br>±1.0 % FS | General pressure difference measurements between two points |